%% file: mimo_paper.tex
\PassOptionsToPackage{prologue,dvipsnames}{xcolor}
\documentclass[]{bytedance_seed}

\usepackage[toc,page,header]{appendix}

\usepackage{minitoc}

\usepackage{amsfonts}
\usepackage{amsmath}
\usepackage{amssymb}
\usepackage{lineno}
\usepackage{wrapfig}  
\usepackage{booktabs} 
\usepackage[bottom]{footmisc}

\usepackage{CJKutf8}
\usepackage{setspace}
\usepackage{arydshln}
\usepackage{xspace} 

\usepackage{graphicx}
\usepackage{subcaption}
\usepackage{colortbl}

\usepackage{xcolor}

\definecolor{linkcolor}{rgb}{0.956,0.298,0.235}
\definecolor{citecolor}{HTML}{1976D2}
\hypersetup{
    colorlinks=true,    
    linkcolor=citecolor,     
    filecolor=magenta,  
    urlcolor=cyan,      
    citecolor=citecolor,    
}

\newcommand{\ours}{Mogao}

\newcommand{\eg}{e.g.\xspace}
\newcommand{\ie}{i.e.\xspace}

\definecolor{customblue}{HTML}{DAE8FC}
\definecolor{customyellow}{HTML}{FFF2CC}

\title{\ours: An Omni Foundation Model\\for Interleaved Multi-Modal Generation}
\author[*]{Chao Liao}
\author[*]{Liyang Liu}
\author[*]{Xun Wang}
\author[*]{Zhengxiong Luo}
\author{Xinyu Zhang} 
\author{Wenliang Zhao} 
\author{Jie Wu}
\author{Liang Li}
\author[\dagger]{Zhi Tian}
\author[\ddagger]{Weilin Huang}

\affiliation{ByteDance Seed}

\contribution[*]{Equal Contribution}
\contribution[\dagger]{Project Lead}
\contribution[\ddagger]{Corresponding authors}

\abstract{
Recent progress in unified models for image understanding and generation has been impressive, yet most approaches remain limited to single-modal generation conditioned on multiple modalities. In this paper, we present {\ours}, a unified framework that advances this paradigm by enabling interleaved multi-modal generation through a causal approach.
{\ours} integrates a set of key technical improvements in architecture design, including a deep-fusion design, dual vision encoders, interleaved rotary position embeddings, and multi-modal classifier-free guidance, which allow it to harness the strengths of both autoregressive models for text generation and diffusion models for high-quality image synthesis. 
These practical improvements also make {\ours} particularly effective to process interleaved sequences of text and images arbitrarily.
To further unlock the potential of unified models, we introduce an efficient training strategy on a large-scale, in-house dataset specifically curated for joint text and image generation.
Extensive experiments show that {\ours} not only achieves state-of-the-art performance in multi-modal understanding and text-to-image generation, but also excels in producing high-quality, coherent interleaved outputs. Its emergent capabilities in zero-shot image editing and compositional generation highlight {\ours} as a practical omni-modal foundation model, paving the way for future development and scaling the unified multi-modal systems.
}

\date{\today}

\begin{document}
\maketitle



\begin{figure}[pt]
\begin{center}
\includegraphics[width=1.\linewidth]{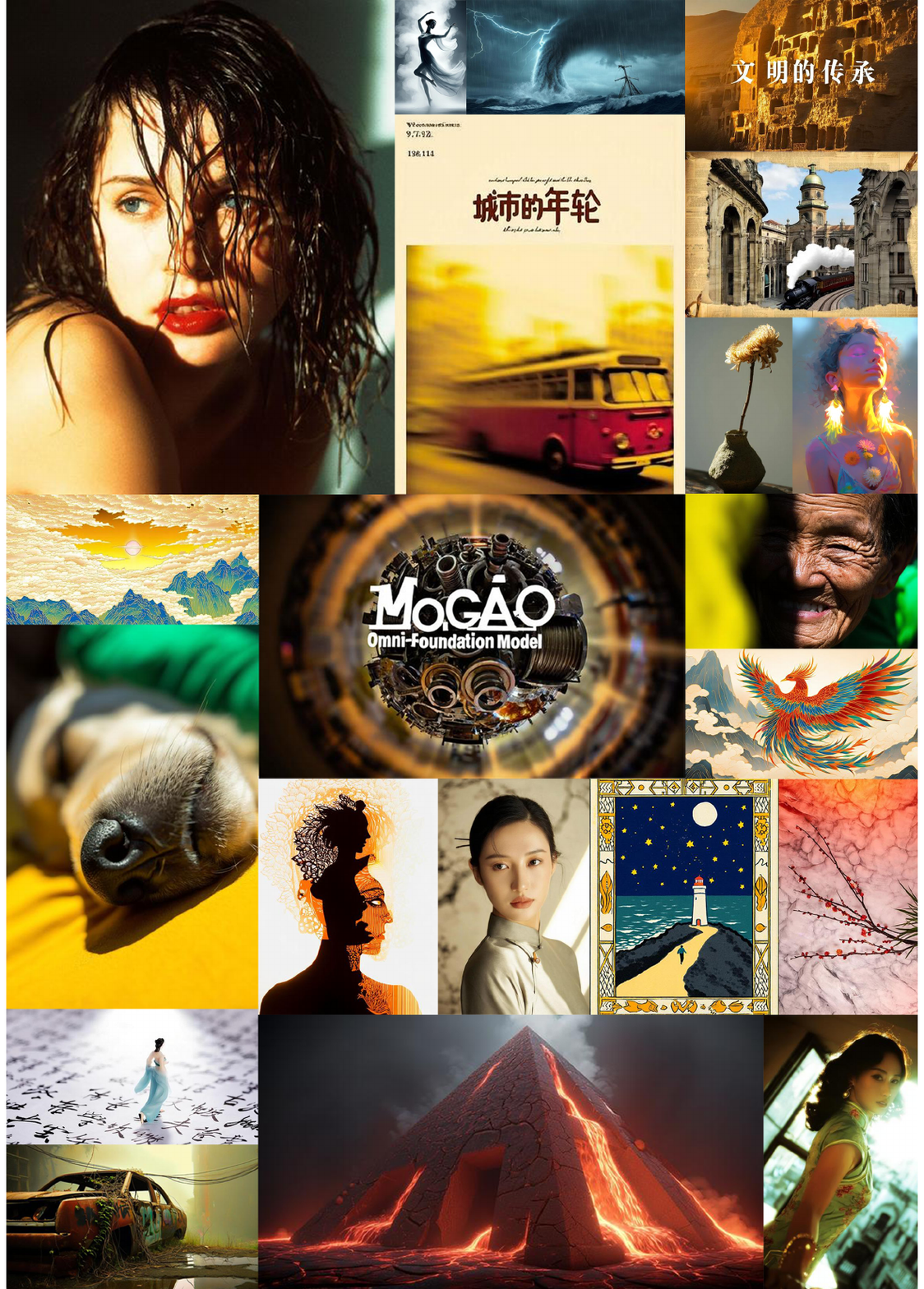}
\end{center}
\label{fig:teaser}
\vspace{-1pt}
\caption{Mogao Text-to-Image visualization.}
\end{figure}

\newpage

 \begin{figure*}[!h]
 \centering
 \includegraphics[width=1.\linewidth]{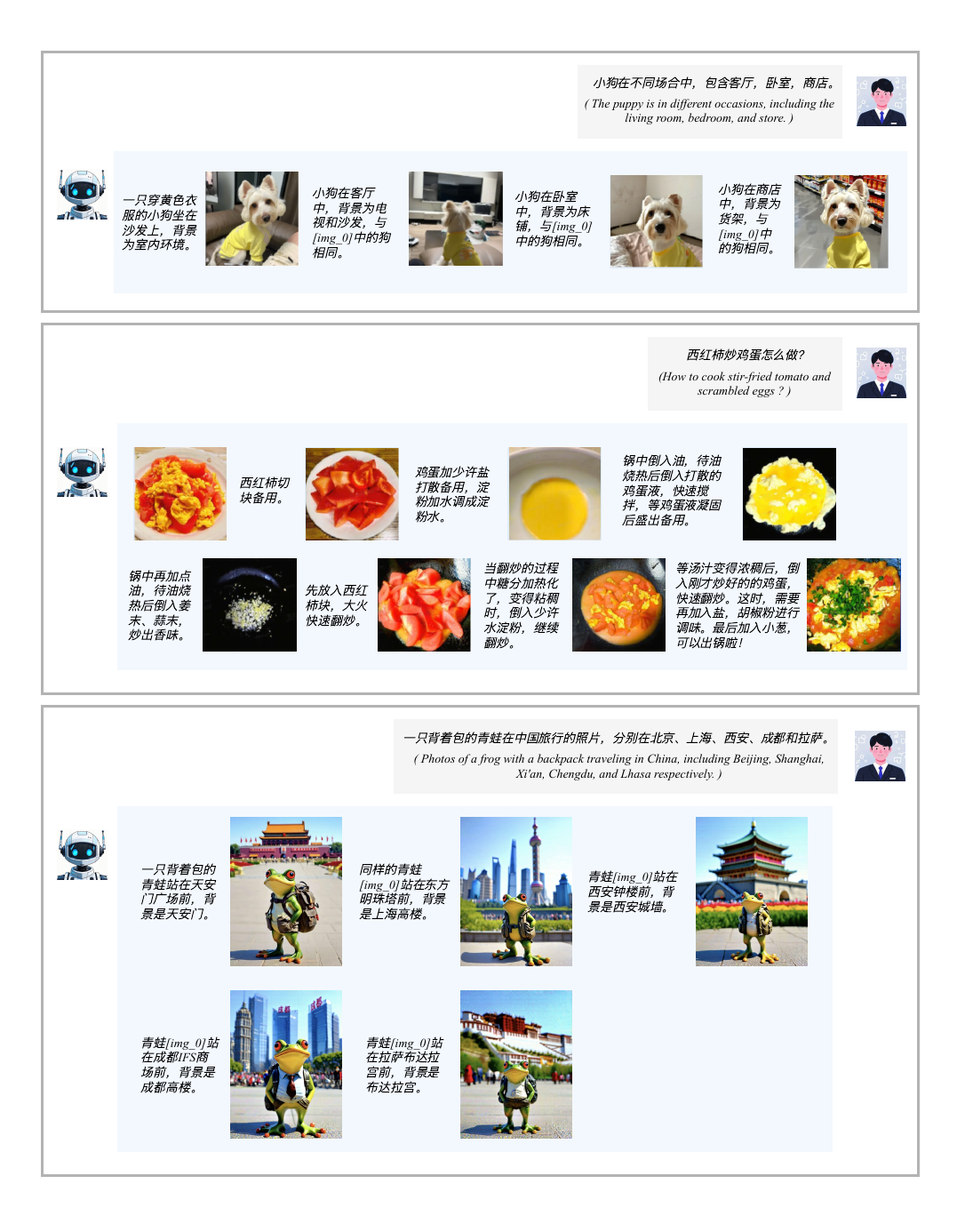}
 \caption{{\ours} Interleaved Multi-Modal Generation Visualization.}
 \label{fig:interleave_vis}
 \end{figure*}

\input{mimo_sections/introduction}
\input{mimo_sections/relatedwork}
\input{mimo_sections/approach}
\input{mimo_sections/experiments}
\input{mimo_sections/conclusion}

\clearpage

\bibliographystyle{plainnat}
\bibliography{mimo_main}

\clearpage



\end{document}

%% file: mimo_sections/introduction.tex
\section{Introduction}
Generative models have shown significant potential to advance towards artificial general intelligence (AGI). Contemporary models, such as large language models (LLMs)~\cite{gpt3,llama} and text-to-image (T2I) models~\cite{ldm,flux2023,wang2025simplear,gong2025seedream20nativechineseenglish,gao2025seedream3}, typically process single-modal inputs and generate single-modal outputs. However, to interact with users in a human-like manner, models must be capable of interpreting arbitrarily interleaved multi-modal inputs and generating corresponding multi-modal outputs. Such models can seamlessly integrate multiple modalities, as illustrated in Fig.~\ref{fig:interleave_vis}, to convey meaningful and more accurate information through a combination of text and visuals.

Early efforts in unified multi-modal modeling have leveraged auto-regressive (AR) models for text generation. By introducing visual tokenizers and de-tokenizers, image patches can be processed as text tokens, facilitating the seamless extension of traditional LLMs to large multi-modal models. These models have demonstrated excellent performance in visual understanding~\cite{qwen2.5-vl,openai2024gpt4ocard}. However, AR-based approaches for image generation often encounter limitations, including low training efficiency~\cite{rar}, exposure bias~\cite{han2024infinity}, and suboptimal image tokenization~\cite{mar}. Inspired by the success of diffusion-based models in visual generation~\cite{ddpm,flux2023}, recent research has explored hybrid AR-diffusion architectures to advance unified multi-modal modeling. For instance, Transfusion~\cite{zhou2024transfusion} integrates an AR model and a diffusion model by sharing transformer weights, enabling text generation via next-token prediction and image generation via iterative denoising. While these approaches leverage the strengths of both paradigms, the implications of shared weights remain underexplored.

In this work, we develop a practically efficient hybrid architecture that seamlessly integrates AR-based text generation with diffusion-based image generation. Text tokens and image embeddings are combined into a unified input sequence, processed by a transformer equipped with modality-specific QKV and FFN layers, diverging from modality-shared designs. We introduce a novel position embedding scheme that assigns distinct modality-specific IDs and frequency patterns. Supervision is applied separately: next-token prediction loss for text and diffusion loss for images.
To disentangle understanding and generation tasks, we decouple the Multi-Layer Perceptron (MLP) parameters within the transformer and employ different image encoders—such as a Vision Transformer (ViT) for understanding and a Variational Autoencoder (VAE) for generation. This separation mitigates task interference, enhancing performance across both domains.

In contrast to conventional unified models that depend heavily on paired data supervision, our approach allows \textbf{\ours}\footnote{The Mogao Caves are the best known of the Chinese Buddhist grottoes, which contain some of the finest examples of Buddhist art spanning a period of 1,000 years.} to pioneer native interleaved multi-modal modeling. To facilitate this, we construct a ten-million-scale interleaved multi-modal dataset. Additionally, we design a compute-efficient training strategy that simultaneously optimizes teacher-forced textual tokens and diffusion-based corrupted visual tokens within interleaved sequences. A central innovation lies in our multi-conditioned image generation paradigm: unlike traditional models constrained to single-modal inputs (e.g., text-to-image), our method dynamically incorporates contextual information from both preceding text and images. Moreover, a standard classifier-free guidance (CFG) approach risks image repetition during interleaved generation due to varying transferability between intra- and cross-modal tokens. To mitigate this, we introduce a dual CFG mechanism that integrates both empty and visual-only conditions, thereby enhancing the precision of image generation.

The primary contributions of \textbf{\ours} encompass the following three aspects:
\begin{itemize}
    \item We propose a conceptually elegant and highly effective unified architecture that seamlessly integrates autoregressive and diffusion models, enabling superior both text and image generation.
    \item We introduce a suite of innovative technical improvements that facilitate efficient training and inference for interleaved multi-modal understanding and generation.
    \item We conduct comprehensive experiments, including ablation studies, to rigorously validate our approach, achieving state-of-the-art performance across multiple open benchmarks, substantiated by human evaluations for multi-modal generation.
\end{itemize}

%% file: mimo_sections/relatedwork.tex
\section{Related Work}

Visual understanding and generation have traditionally been handled by separate models: vision language models~\cite{llava,liu2024improved} for understanding and diffusion models~\cite{ddpm,ldm,gao2025seedream3,gong2025seedream20nativechineseenglish} for generation. Recently, efforts to unify the two tasks into a single foundation model have gained momentum, driven by the potential to produce higher quality, more consistent multimodal content~\citep{flamingo, liquid}.

\paragraph{\textbf{Unified Framework}}
A central challenge in building unified models is the misalignment between text and visual modalities. State-of-the-art vision-language models (VLMs) are typically built on GPT-style large language models (LLMs)~\cite{gpt1,gpt2,gpt3}, and trained via next-token prediction~\cite{wang2024qwen2,qwen2.5-vl,llava-onevision}. 
In contrast, high-performance visual generation models are based on diffusion models~\cite{flux2023,esser2024scaling}. Recent efforts attempted to unify visual generation within an auto-regressive (AR) framework by using next-token prediction. For example, the Emu series~\cite{sun2023emu,emu2} consider images as continuous feature vectors concatenated with text embeddings, allowing a language model~\cite{llama} to generate both text tokens and image vectors autoregressively. Similarly, Chameleon~\cite{team2024chameleon} replaces continuous image features with discrete visual tokens. However, AR methods often struggle to generate high-quality images~\cite{maskgit, mar, var}, compared to diffusion-based approaches.

Other approaches embed diffusion modules directly into an LLM backbone. For example, recent approaches such as Show-O~\cite{xie2024show} and TransFusion~\cite{zhou2024transfusion}, are able to generate text token-by-token, and synthesize an images through a diffusion process, by sharing a transformer backbone~\cite{peebles2023scalable, transformer}. This can improve the performance of visual generation, but might sacrifice the understanding capability, due to the conflicts of two tasks learning with shared parameters~\cite{lin2024moma,liang2024mixture}. In \textbf{\ours}, we adopt the MMDiT architecture~\cite{esser2024scaling}, which decouples two tasks by using separate text and visual parameters to reduce such cross-modal conflicts.

\paragraph{\textbf{Unified Representation}}
Another challenge lies in the differing requirements of visual representations for understanding and generation~\cite{tokenflow,unitok}. Features optimized for understanding may be suboptimal for generation, making visual tokenizer design critical. For example, SEED~\cite{seed,seed2} supervises its tokenizer using a reconstruction loss for generation while aligning tokenized features with CLIP for understanding~\cite{radford2021learning}. Recent methods~\cite{unitok,vila-u} adopt a similar strategy, aligning the latent space of VQGAN~\cite{vqgan} or VAE~\cite{vae,ldm} with semantic encoders~\cite{zhai2023sigmoid,oquab2023dinov2}. Although this improves overall performance, it still involves trade-offs between understanding and generation. JanusFlow~\cite{ma2024janusflow} mitigates this by employing separate tokenizers: SigLIP~\cite{zhai2023sigmoid} for understanding and a VAE for generation, enhancing both tasks simultaneously.

\paragraph{\textbf{Interleaved Multi-Modal Generation}}
A key application of omni-foundation models is interleaved multi-modal generation. Prior work~\cite{emu2,cm3leon} shows that AR models trained on interleaved data can generate both text and images conditioned on mixed modalities, enabling unified capabilities such as image understanding, text-to-image generation, image editing, and composition. Further developments~\cite{zhou2024transfusion,omnigen} enhance visual fidelity by replacing token-wise synthesis with diffusion-based iterative denoising. However, these models typically focus on single-turn outputs, generating either text or image in isolation. In \textbf{\ours}, we extend this to multi-turn interleaved generation.

%% file: mimo_sections/approach.tex
\section{\ours}
\label{sec:method}

Vision Language Models (VLMs)~\cite{liu2024improved, lu2024deepseek, wang2024qwen2} perform visual understanding by modeling the conditional distribution $P\left(\mathbf{x}_{\text{txt}}\mid\mathbf{x}_{\text{img}}, \mathbf{x}_{\text{txt}}\right)$, while Text-to-Image (T2I) models~\cite{flux2023, esser2024scaling} perform image generation using $P\left(\mathbf{x}_{\text{img}}\mid\mathbf{x}_{\text{txt}}\right)$. \textbf{\ours} uses a unified backbone to model the joint distributions of images and text, naturally possessing the ability of both understanding and image generation. In addition, in this paper, we extend its generative capabilities by using a causal approach, which enables \textbf{\ours} to implement a more complicated task: interleaved multi-modal generation, whereby using any modalities information generated in the past as a condition to generate new outputs in any modalities.

\paragraph{\textbf{Multi-Modal Understanding}}
 Text data can be represented by the discrete token sequence $
 \mathbf{z}=\left(z_{0},\dots,z_{L}\right)$, and LLMs model its joint distribution in a causal manner. Based on LLM, VLM converts visual features into continuous token sequences by integrating visual encoders~\cite{zhai2023sigmoid, radford2021learning} to process visual input. By training on large-scale multi-modal datasets, VLM expands the conditional distribution:
 
 
 
 \begin{equation}
 \label{eq:vlm}
 \log {P}_{\mathbf{\theta}_\text{VLM}}\left(\mathbf{z}\mid\mathbf{x}\right)=\sum_{l=0}^{L}\log {P}_{\mathbf{\theta}_\text{VLM}}\left(z_{l+1}\mid z_{0},\dots,z_{l},\mathbf{x}\right) 
 \end{equation}

 Where $\mathbf{\theta}_{\text{VLM}}$ and $\mathbf{x}=\left(x_{0},\dots,x_{N}\right)$ denote the parameters of VLM and the continuous token sequence of visual input respectively, and $\mathbf{x}$ can be $\varnothing$. In general, both LLMs and VLMs only output text tokens $z_i$. Therefore, we can optimize the model by minimizing the cross-entropy between $P_\theta$ and the data distribution, which is the so-called next token prediction (NTP) loss:

 \begin{equation}
 \label{eq:lm_loss}
 \mathcal{L}_{\text{NTP}} = \mathbb{E}_{z_{i}} \left[ -\log P_{\theta_\text{VLM}}\left( z_{l+1}\mid z_{0},\dots,z_{l},\mathbf{x}\right) \right]
 \end{equation}

\paragraph{\textbf{Visual Generation}}
 In this paper, we adopt rectified flow matching~\cite{liu2022flow} to model image distribution, which is simple in concept but performs well in text-to-image generation~\cite{esser2024scaling}. Given the image distribution $\mathbf{x}\sim\pi$ and a simple prior distribution $\boldsymbol{\epsilon}\sim\mathcal{N}\left(0,1\right)$, we construct a linear trajectory defined over time $t$:
 
 \begin{equation}
 \label{eq:flow_fwd}
 \mathbf{x}_t = t\cdot\mathbf{x}+\left(1-t\right)\cdot\boldsymbol{\epsilon}, \; t\in\left[0,1\right]
 \end{equation}

 The following ordinary differential equation (ODE) can be derived from Eq. (\ref{eq:flow_fwd}):

 \begin{equation}
 \label{eq:flow_ode}
 \frac{\mathrm{d}\mathbf{x}_t}{\mathrm{d}t} = \mathbf{v}\left(\mathbf{x}_t,t\right)
 \end{equation}

 where the velocity field $\mathbf{v} = \mathbf{x} - \boldsymbol{\epsilon}$. To generate an image, we can perform the reverse time integration of ODE in Eq. (\ref{eq:flow_ode}) starting from $\mathbf{x}_{0}\sim\mathcal{N}\left(0,1\right)$, specifically, $\mathbf{x}_1 = \int_{0}^{1} \mathbf{v}(\mathbf{x}_t, t)\mathrm{d}t$. Therefore, for image generation, we need to train the network parameterized by $\mathbf{\theta}$ to approximate $\mathbf{v}$ by optimizing the flow matching loss:
 
 \begin{equation}
 \label{eq:flow_loss}
 \mathcal{L}_{\text{flow}} = \mathbb{E}_{t,\mathbf{x}\sim\pi,\boldsymbol{\epsilon} \sim \mathcal{N}\left(0,1\right)}\|\mathbf{v}_{\mathbf{\theta}}(\mathbf{x}_t, t) - (\mathbf{x} - \boldsymbol{\epsilon}) \|_2^2
 \end{equation} 
 
\paragraph{\textbf{Interleaved Multi-Modal Generation}}

 In contrast to contemporary models which unify visual understanding and generation~\cite{zhou2024transfusion, xie2024show, wang2024emu3, ma2024janusflow}, {\ours} modeling the sequential distributions at the \textbf{modal level} in a causal manner:
 
 \begin{equation}
 \label{eq:il_modeling}
 \log {P}_{\mathbf{\theta}}\left(\mathbf{x}\right)=\sum_{t=0}^{T}\log {P}_{\mathbf{\theta}}\left(\mathbf{x}_{t+1}\mid\mathbf{x}_{0},\dots,\mathbf{x}_{t}\right),\;\mathbf{x}_t \in \{\mathbf{x}_{\text{img}},\mathbf{x}_{\text{txt}}\}
 \end{equation}

 Here, $\mathbf{\theta}$ denotes the parameters of the model; notably, $\mathbf{x}_t$ represents visual or text modality rather than an individual token. Once $\mathbf{x}_{t+1}$ is generated, it will be added to the historical information as the condition. Thus, {\ours} can facilitate the generation of any modality conditioned on an arbitrary mixture of modalities, to wit, interleaved multi-modal generation. 

 For the interleaved multi-modal generation task, we need to calculate the NTP loss in Eq.~\ref{eq:lm_loss} for text and flow matching loss in Eq.~\ref{eq:flow_loss} for noised image simultaneously in a sample of interleaved images and texts. Therefore, the overall training loss is a weighted combination of these two losses:
 
 \begin{equation}
 \label{eq:combine_loss}
 \mathcal{L}_{\text{total}} = \lambda \cdot \mathcal{L}_{\text{NTP}} + \mathcal{L}_{\text{flow}}
 \end{equation}

  It is worth noting that during training, {\ours} uses two special tokens to distinguish between visual and textual modalities. All image sequences will have the tokens \texttt{<|vision\_start|>} and \texttt{<|vision\_end|>} inserted at the beginning and end, respectively. During inference, Mogao decides whether to generate an image based on the context. When generating \texttt{<|vision\_start|>}, it switches to image generation mode, and when generating \texttt{<|vision\_end|>}, it switches to text generation mode.

 \begin{figure*}[h]
 \centering
 \includegraphics[width=1.0\linewidth]{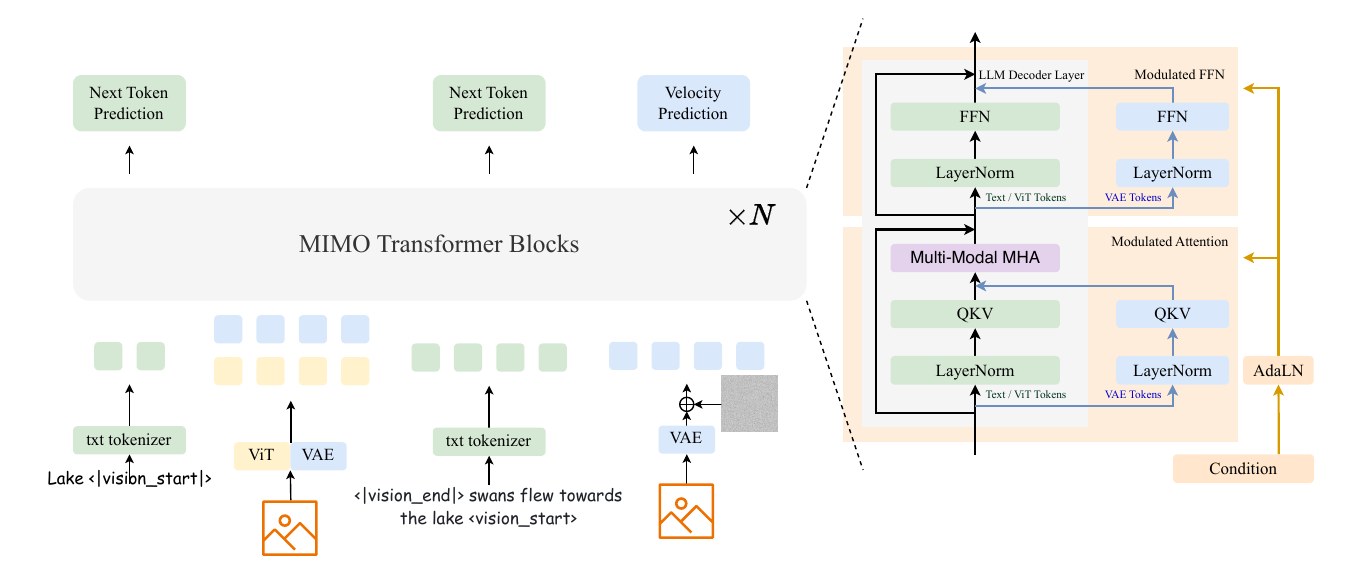}
 \caption{\textbf{Overall framework of the proposed \ours.} The model consists of $N$ modified LLM decoder layers, using different QKV matrix and FFN to process visual tokens. When the image is used as a condition, VAE and ViT are both used to extract visual features, and VAE is used alone when generating images. It is worth noting that we assign the same position id to the ViT token and its corresponding VAE token. The condition will modulate the visual token through an AdaLN layer.}
 \label{fig:mogao}
 \end{figure*}
 
\subsection{Model}
\label{sec:method_model}
The detailed architecture of {\ours} is illustrated in Fig.~\ref{fig:mogao}.

\paragraph{\textbf{Improved Dual Visual Encoder for Interleaved Multi-Modal Generation}}
  For visual generation, current mainstream T2I models use the Variational Autoencoder (VAE) to encode images into a low-dimensional latent space before performing the diffusion process. However, we find that compared to the pre-aligned semantic visual representations obtained from ViT (\eg, SigLIP~\cite{zhai2023sigmoid}) in VLM models, the visual representations produced by VAE are inadequate for visual understanding, which is also mentioned in works such as JanusFlow~\cite{ma2024janusflow}. JanusFlow proposed utilizing a VAE as the visual encoder during visual generation (where images serve as the target distribution) while employing a ViT as the visual encoder during visual understanding (where images act as the condition). 
 

  In the context of {\ours}'s interleaved multi-modal generation, we further refined this strategy. Specifically, when an image serves as a condition, {\ours} simultaneously extracts both ViT and VAE visual representations and appends them into the historical sequence. 
  For multi-modal understanding tasks, i.e., text token generation, text tokens will only attend to ViT tokens and text tokens in the historical sequence. 
  Conversely, for multi-modal generation tasks, i.e., image generation, noisy VAE tokens attend to all tokens in the historical sequence, including ViT, VAE, and text tokens. 
  Because visual representations from ViT can provide a robust semantic representation of images, facilitating improved contextual alignment for image generation.

\paragraph{\textbf{Deep-Fusion Architecture}}
  Our model architecture is constructed upon a pre-trained large language model (LLM) and incorporates some design inspired by MMDiT~\cite{esser2024scaling} to enhance visual generation capabilities. Specifically, we employ Qwen2.5~\cite{yang2024qwen2} as the backbone and modify each transformer block as follows: First, we use a unified self-attention layer to concurrently process both visual and text sequences. 
  Secondly, considering the disparities between visual and text modalities, we employ distinct multilayer perceptrons (MLPs) in the feed-forward network (FFN) and linear projectors in the attention blocks to process them separately. Notably, as shown in Fig.~\ref{fig:mogao}, since ViT provides visual representations with high-level semantics aligned with text, we route ViT tokens to the text branch.
  Finally, since the rectified flow~\cite{liu2022flow} necessitates predicting the velocity at the current timestep, the timestep embedding will modulate the visual feature maps of each block via an AdaLN layer~\cite{peebles2023scalable}.
  
  Building upon the aforementioned design, we observe that this architecture can achieve high-quality image generation outcomes without relying on the text encoder (\eg, CLIP~\cite{radford2021learning} or T5~\cite{raffel2020exploring}) typically required in conventional T2I models~\cite{flux2023,li2024hunyuan, esser2024scaling}. We attribute this to the deep fusion of the LLM's text representations and image features at each layer. Furthermore, this design enables the model to naturally inherit the long-context capabilities of the LLM, which is critical for interleaved multi-modal generation over extended sequences.

\paragraph{\textbf{Interleaved Rotary Position Embedding}}

  Both the input and output of \textbf{\ours} are multi-modal, so the model needs to capture three-dimensional positional relationships: $H$ and $W$ are used to represent the spatial positional information of the image, and $T$ is used for the temporal positional relationship between all images and text tokens. 
  Rotary Positional Embedding (RoPE)~\cite{su2024roformer} is a widely used and effective scheme. We improved on it and proposed \textbf{I}nter\textbf{l}eaved RoPE (\textbf{IL-RoPE}). Different from many recent works on 3D-RoPE~\cite{qwen2.5-vl, agrawal2024pixtral, wei2025videorope}, we have improved the frequency allocation of each dimension and the setting of position ID for the scenario of simultaneously generating text and images.


 \begin{figure}[h]
    \centering
    \begin{subfigure}[b]{0.725\textwidth}
        \includegraphics[width=\textwidth]{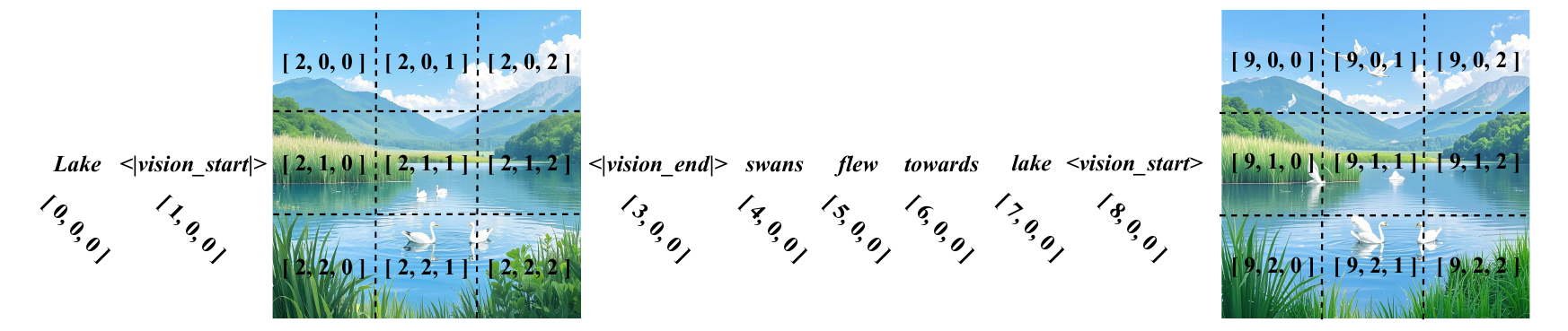}
        \subcaption{The setting of position ID.}
        \label{fig:il_rope_posid}
    \end{subfigure}
    \begin{subfigure}[b]{0.225\textwidth}
        \includegraphics[width=\textwidth]{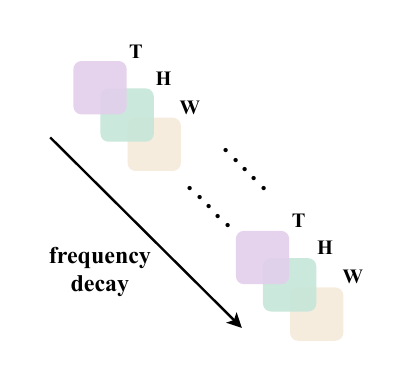}
        \subcaption{Frequency allocation.}
        \label{fig:il_rope_freq}
    \end{subfigure}
    \caption{Interleaved 3d-RoPE.}
    \label{fig:il_rope}
\end{figure}

  As shown in Fig.~\ref{fig:il_rope_posid}, in terms of the design of position ID, for images, the spatial position ID of ${H}/${W} of each image is calculated from 0, so that all images are guaranteed to have consistent position information in space, which is conducive to image editing, multi-image generation, etc., because these tasks need to refer to previous images for generation. Meanwhile, the temporal position ID will increase for each token. Note that the temporal position ID of all tokens in an image is consistent. 
 
  In terms of frequency allocation, M-RoPE~\cite{wang2024qwen2} proposes assigning higher frequencies to $T$, intermediate and lower frequencies to $H$ and $W$ respectively within the $d=128$ dimensions:
 \begin{equation}
 \label{eq:mrope}
 \Theta_{T}=\left\{\beta^{-\frac{2i}{d}}\mid i\in[0, 16)\right\};\;\Theta_{H}=\left\{\beta^{-\frac{2i}{d}}\mid i\in[16, 40)\right\};\;\Theta_{W}=\left\{\beta^{-\frac{2i}{d}}\mid i\in[40, 64)\right\}
 \end{equation}
  where $\Theta$ denotes the rotation frequency applied to a specific dimension, and $\beta$ denotes the base frequency. Such design results in the temporal dimension $T$ focusing on capturing local relationships, while the spatial dimensions $H$ and $W$ prioritize long-range dependencies. This significantly degrades image generation performance, as image synthesis heavily relies on local semantic information; for instance, a 256-resolution image comprises only 16 tokens along $H$ and $W$ dimensions respectively~\cite{flux2023}. Inspired by the above observation, IL-RoPE interleaves frequency assignment for $\{T, H, W\}$ within the $d$ dimension to ensure that each dimension can capture long-range and local semantic information in a balanced manner, as demonstrated in Fig.~\ref{fig:il_rope_freq}.
 \begin{equation}
 \label{eq:il_rope}
 \Theta_{T}=\left\{\beta^{-\frac{2i}{d}}\mid i=3j\right\},\;\Theta_{H}=\left\{\beta^{-\frac{2i}{d}}\mid i=3j+1\right\},\;\Theta_{W}=\left\{\beta^{-\frac{2i}{d}}\mid i=3j+2\right\}
 \end{equation}
  As shown in Eq. (\ref{eq:il_rope}), we assign the first $48$ channels to $T/H/W$ dimension alternately, and then assign the left $16$ channels to $T$ dimension for modeling long sequences.

\paragraph{\textbf{Multi-modal Classifier-free Guidance}}
  Following previous works, we use Classifier-free guidance (CFG)~\cite{ho2022classifier} to enhance generation quality, which combines conditional predictions with unconditional predictions to yield results that more closely align with the specified conditions:
 
 \begin{equation}
 \label{eq:t2i_cfg}
 \nabla_\mathbf{x}\log p\left(\mathbf{x}\mid\mathbf{c}\right)=\gamma\left(\nabla_\mathbf{x}\log p\left(\mathbf{x}\mid\mathbf{c}\right)-\nabla_\mathbf{x}\log p\left(\mathbf{x}\right)\right)+\nabla_\mathbf{x}\log p\left(\mathbf{x}\right)
 \end{equation}
 
  where $\mathbf{c}$ and $\gamma$ denote the condition and CFG coefficient respectively. The condition of T2I models consists only of the text modality $\mathbf{c}_\text{txt}$, but in the interleaved multi-modal generation scenario presented in this paper, condition $\mathbf{c}$ can be decomposed into two modalities $\mathbf{c}_\text{txt}$ and $\mathbf{c}_\text{img}$. In practice, we observe that the image within the condition serves as a shortcut for the model in generating the subsequent image, resulting in a phenomenon akin to temporal stagnation~\cite{lin2024stiv}.
 Inspired by~\cite{brooks2023instructpix2pix}, we introduce distinct CFG coefficients $\gamma$ and $\gamma_\text{img}$ to control the influence of different modalities:
 
\begin{equation} \label{eq:mogao_cfg}
\begin{split}
\nabla_\mathbf{x}\log p\left(\mathbf{x}\mid\mathbf{c}_{\text{img}},\mathbf{c}_{\text{txt}}\right) =\;& \gamma\left(\nabla_\mathbf{x}\log p\left(\mathbf{x}\mid \mathbf{c}_{\text{img}},\mathbf{c}_{\text{txt}}\right)-\nabla_\mathbf{x}\log p\left(\mathbf{x}\mid\mathbf{c}_{\text{img}}\right)\right)+\\
&\gamma_{\text{img}}\left(\nabla_\mathbf{x}\log p\left(\mathbf{x}\mid \mathbf{c}_{\text{img}}\right)-\nabla_\mathbf{x}\log p\left(\mathbf{x}\right)\right)+\nabla_\mathbf{x}\log p\left(\mathbf{x}\right)
\end{split}
\end{equation}

  Note that when $\gamma_\text{img}=\gamma$, Eq. (\ref{eq:mogao_cfg}) degrades to Eq. (\ref{eq:t2i_cfg}), where the frequently used cross-modal CFG coefficient may be too large for the intra-modal condition. This enables us to effectively regulate the influence of different modalities on image generation. We use the CFG coefficient $\gamma=7.5$ and $\gamma_\text{img}=1.5$ respectively.

\subsection{Data}
  We use a mixture of text-only, visual understanding, image generation, and interleaved multi-modal data to train our model. The text-only and visual understanding data are inherited from DouBao LM and VLM datasets. We leverage the image generation data used to train SeedDream~\cite{gong2025seedream20nativechineseenglish} for improving image quality and diversity. As for multimodal interleaved data, we curate it from publicly available websites and videos. For data with native texts and images, we keep the original item order during training. For video clips, we train a vision-language model to get the caption of each frame sampled from the clip. The training sample is composed of sampled frames and captions in an intertwined manner.

\subsection{Training}
\label{sec:method_training}

\paragraph{\textbf{Native Resolution}}
  The strategy of resizing images to a fixed resolution is evidently suboptimal, as it simultaneously reduces the efficiency of model training and inference while substantially limiting the model's visual understanding capabilities~\cite{lee2023pix2struct, dehghani2023patch, wang2024qwen2}. Consequently, throughout the entire training phase, we adopt a native resolution strategy, converting input images into a variable number of visual tokens while preserving their aspect ratios. Furthermore, to bridge image representations between visual generation and understanding, we utilize this strategy consistently across both tasks.

\paragraph{\textbf{Global Batch Reduced Loss}}
  When training with mixed-source multi-modal data on multiple GPUs, we need to carefully balance losses of tokens with variable weights to achieve a good trade-off among tasks. Suppose on $n$-th GPU, the $i$-th token has loss $\mathcal{L}_i^n$ with weight $w_i^n$, it is ideal to take tokens on all GPUs as a global batch and compute the gradient with respect to model parameter $W$ by $\frac{\partial}{\partial W}\left(\frac{\sum_n\sum_{i}w_i^n\mathcal{L}_i^n}{\sum_n\sum_iw_i^n}\right)$. However, it requires gathering the weights and losses of tokens on all ranks, which will definitely harm training efficiency due to the large amount of communication cost. An approximate method used frequently is to compute per-rank loss and average gradient among GPUs, \ie, $\frac{1}{N}\sum_n\frac{\partial}{\partial W}\left(\frac{\sum_{i}w_i^n\mathcal{L}_i^n}{\sum_iw_i^n}\right)$. However, the gradient computed by this strategy can be biased from the ideal one. We propose to use $\frac{1}{N}\sum_n\frac{\partial}{\partial W}\left(\frac{\sum_{i}w_i^n\mathcal{L}_i^n}{\frac{1}{N}\sum_n\sum_iw_i^n}\right)$ as a proxy, which can be easily derived to be equal to the unbiased gradient. Specifically, we first average-reduce the sum of weights on each rank and use the reduced weight as the normalization term. The communication cost of $N$ scalars is negligible and will not introduce any training inefficiency since we apply async-reduce in practice.

\paragraph{\textbf{Efficient Complete Teacher Forcing}}
  In interleaved multi-modal generation where interleaved samples consist of visual and textual modalities, a critical challenge arises from the discrepancy between training and inference since all visual elements are designated to be predicted during training. Specifically, while images undergo diffusion forward processes with multi-level Gaussian noise perturbations, subsequent modalities (text or images) are trained under these noisy conditions. This introduces a domain shift between the perturbed training phase and the clean inference phase.


  \begin{figure}[h]
    \centering
    \begin{subfigure}[b]{0.4\textwidth}
        \includegraphics[width=\textwidth]{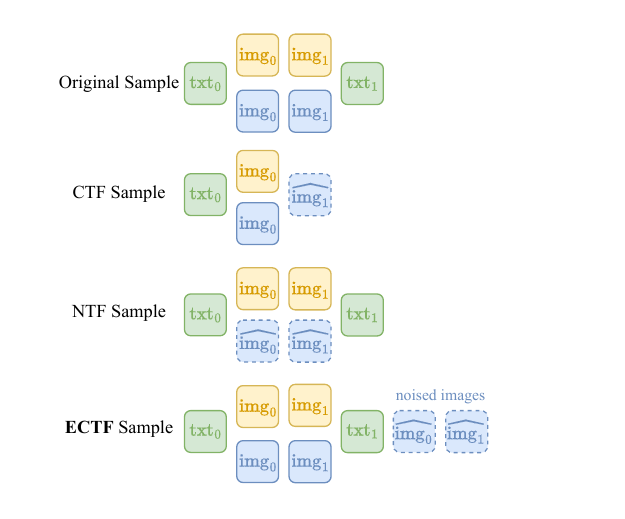}
        \subcaption{Comparison of Samples.}
        \label{fig:sample_comp}
    \end{subfigure}
    \begin{subfigure}[b]{0.45\textwidth}
        \includegraphics[width=\textwidth]{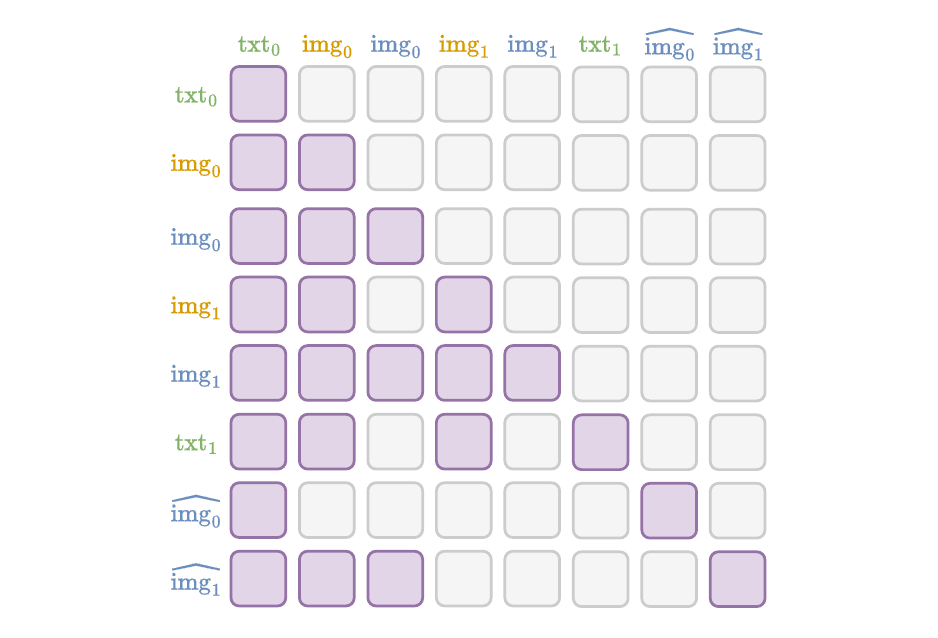}
        \subcaption{Attention Mask of ECTF.}
        \label{fig:ectf_mask}
    \end{subfigure}
    \caption{\textbf{Efficient Complete Teacher Forcing.} The VAE and ViT visual representations of an image are highlighted in \colorbox{customblue}{blue} and \colorbox{customyellow}{yellow} respectively. $\widehat{\text{img}}$ indicates the noisy image. \textbf{a)} As depicted in \ref{sec:method_model}, VAE and ViT token will share the same position id. \textbf{b)} VAE tokens can attend to all tokens, but text tokens and ViT tokens can not attend VAE tokens.}
    \label{fig:ectf}
\end{figure}
 

  While Transfusion~\cite{zhou2024transfusion} mitigates the training-inference discrepancy via constrained noise scales, it fundamentally preserves the mismatch between iterative denoising (training) and single-pass generation (inference). Alternative methods restrict loss computation to only the final image. Despite the theoretical applicability, they suffer from compute complexity which scales quadratically with the sequence length due to duplicated computation of ahead items, making them computationally prohibitive for long multi-modal sequences.
 
  We propose Efficient Complete Teacher Forcing (ECTF) that achieves an optimal trade-off between computational efficiency and conditional consistency. As illustrated in Fig.~\ref{fig:sample_comp}, ECTF introduces clean-noise decoupling: clean images with interleaved sentences are separated from those noisy images through a dynamic causal attention mask in Fig.~\ref{fig:ectf_mask}, eliminating redundant computation in the last perturbation methods.

  Concurrent work~\cite{zhou2025tamingteacherforcingmasked} introduces a comparable technique to our ECTF framework. However, our approach demonstrates broader versatility by effectively handling interleaved multi-modal inputs, whereas their method remains confined to visual modality processing.


%% file: mimo_sections/experiments.tex
\section{Experiments}
\label{sec:exp}

\subsection{Experimental Setup}
\paragraph{\textbf{Implementation Details}}
We use Qwen2.5-3B as the backbone and modify each transformer block as described in Sec.~\ref{sec:method_model} to adapt to unified training. After transforming it into a deep-fusion architecture, the number of model parameters will double; in addition, the AdaLN-Zero layer will be added to each transformer block. Therefore, the final trainable parameter size of {\ours} is 7B. It is worth noting that the deep-fusion architecture can be considered as a \textbf{fixed-routing MoE}, so its per-token computation is only \textbf{equivalent to a 3.5B model}.
We use VAE from Flux~\cite{flux2023} and ViT from Qwen2-VL~\cite{wang2024qwen2} as the visual encoder, and the parameters of visual encoders are frozen during the entire training. During training, benefitting from the native resolution strategy, all images in all tasks do not require additional padding. Multiple samples within a batch will be packed into a sequence to enhance training efficiency.

\paragraph{\textbf{Training Schemes}}
 The training is divided into three stages, as detailed below:
 \begin{enumerate}
     \item \textbf{Stage 1: Low-resolution unified training.} In the first stage, we conducted unified training using both multi-modal understanding data and text-to-image data. The maximum image resolution is limited to $256 \times 256$ to reduce the training difficulty on the image generation. Since we employed a pre-trained LLM that inherently possesses text generation capabilities, and we required more data to model image distributions for image generation, we adjusted the ratio of text-to-image data to multi-modal understanding data to 4:1. Meanwhile, except the visual encoder, all parameters are trainable.
     \item \textbf{Stage 2: High-resolution unified training.} In this stage, we still use multi-modal understanding data and text-to-image data for training, and keep the data ratio unchanged. At the same time, we adjust the resolution from $256 \times 256$ to $512 \times 512$ to achieve higher quality generation.
     \item \textbf{Stage 3: Multi-modal interleaved training.} After the previous two stages, the model has already demonstrated strong capabilities in multi-modal understanding and text-to-image generation. In the final stage, we incorporate multi-modal interleaved generated data for training. At the same time, we still include some multi-modal understanding and text-to-image datasets to maintain the model's performance in multi-modal understanding and image generation, accounting for 20\% and 40\% respectively. The image resolution is maintained at $512\times512$  and ECTF is used to improve training efficiency.
 \end{enumerate}

\paragraph{\textbf{Evaluation Setup}}
 First, we rigorously evaluate {\ours}’s capabilities in image generation and multi-modal understanding through a series of comprehensive benchmarks. Then, we evaluate {\ours}'s superior interleaved multi-modal generation capabilities through comprehensive and diverse qualitative results.
 
 For multi-modal understanding, the general visual capabilities are evaluated on POPE~\cite{li2023evaluating}, MME~\cite{fu2023mme}, MMBench~\cite{liu2023mmbench}, SEEDBench~\cite{li2023seed}, GQA~\cite{hudson2019gqa}, and the knowledge VQA capabilities are evaluated on MMMU~\cite{yue2024mmmu}. 
 
 For the text-to-image generation, the evaluation is conducted mainly on three popular public benchmarks, \ie, GenEval~\cite{geneval}, DPG-Bench~\cite{dpg-bench}, and GenAI-Bench~\cite{lin2024evaluating}. GenEval is a challenge benchmark designed for image-to-text generation. It uses an end-to-end detector~\cite{mask2former,mmdetection} to evaluate the prompt-generation alignments according to the object categories, numbers, color, position, etc. It contains $553$ short prompts, and we generate $4$ images for every prompt for evaluation. DPG-Bench contains $1065$ dense prompts and evaluates the generated images via a vision-language model~\cite{mplug}. GenAI-Bench contains $871$ basic prompts and $1593$ advanced prompts, designed for evaluation on counting, comparing, logic, etc. The performance is quantified by the VQA (visual-question-answer) score~\cite{flant5}. We report our evaluated results on advanced prompts.

 For interleaved multi-modal generation, we first demonstrate {\ours}'s zero-shot image editing capability, and then we demonstrate its automatic image and text interleaved generation, group image generation, and next frame prediction capabilities after SFT.

\subsection{Interleaved Multi-Modal Generation}
\paragraph{\textbf{Visualization}} Fig.~\ref{fig:interleave_vis} demonstrates {\ours}'s interleaved multi-modal generation capability, from which we can observe that {\ours} exhibits strong ID preservation and instruction-following abilities in the interleaved generation of image and text. We highlight that high-quality and coherent text generation enables {\ours} to produce accurate images based on contextual information.

\begin{figure*}[h]
\centering
\includegraphics[width=1.0\linewidth]{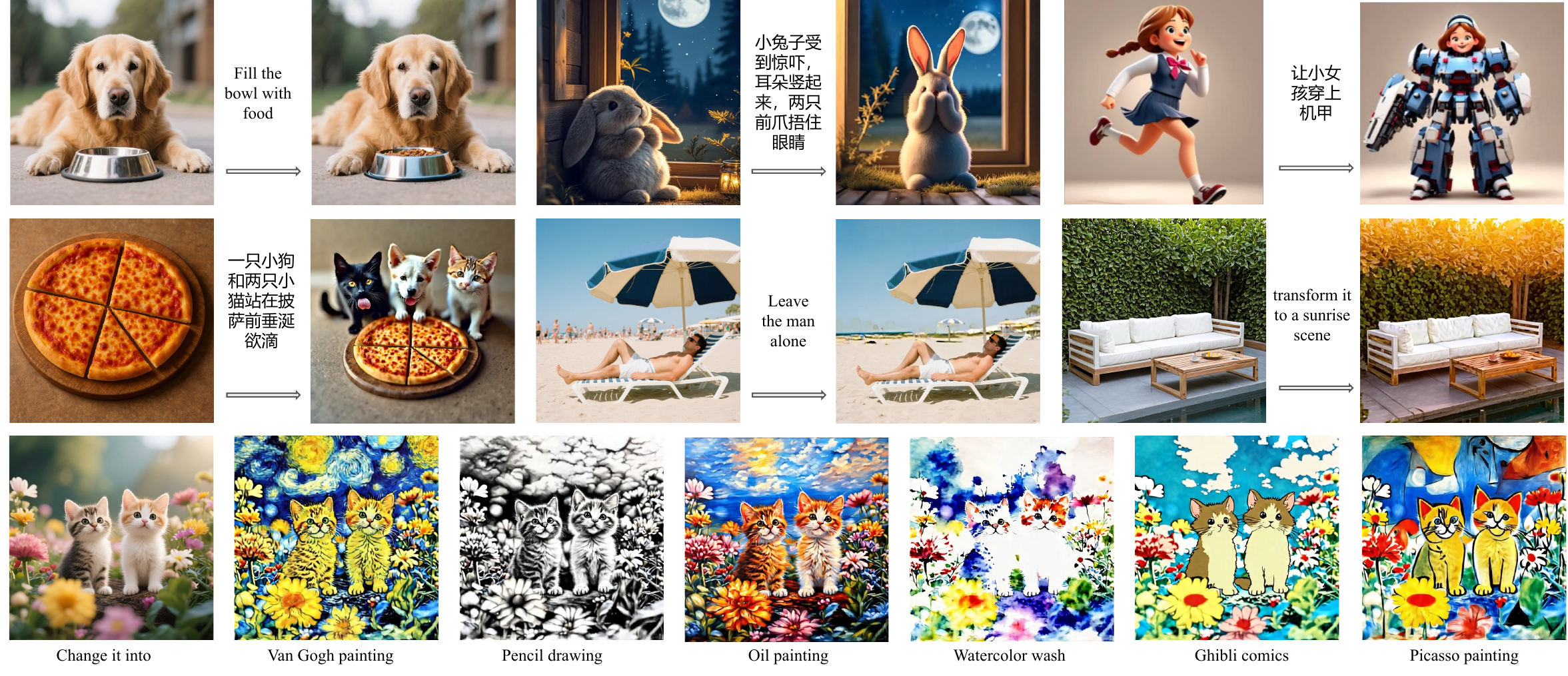}
\caption{\textbf{Mogao Zero-Shot Image Editing Visualization.} The left is the original image for each group.}
\label{fig:edit}
\end{figure*}

\paragraph{\textbf{Zero-Shot Image Editing}} Although {\ours} was trained only on large-scale image-text pair data and interleaved multi-modal data without task-specific training data tailored for image editing, it demonstrates significant potential in zero-shot image editing. As illustrated in Fig.~\ref{fig:edit}, {\ours} can follow instructions and complete tasks such as adding, removing, and modifying elements. It can also respond well to some difficult editing instructions, such as action guidance and perspective and layout modification. {\ours} is currently a pre-trained model without fine-tuning on task-specific data; we emphasize its robust generalization capabilities and the substantial potential revealed through training on interleaved multi-modal data.

\subsection{Text-to-Image Generation}
\input{mimo_sections/table/t2i_eval_table}

\paragraph{\textbf{Benchmark Evaluation}}
 The evaluation results on GenEval~\cite{geneval} are shown in Table~\ref{tab:geneval}. Similar to previous works~\cite{wang2024emu3,janus-pro,nova}, we rewrite the prompts of GenEval to the format of prompts in our training set for a more comprehensive evaluation. As one can see, \textbf{\ours} achieves comparable performance with the previous state-of-the-art unified model, \ie, Janus-Pro-7B~\cite{janus-pro}. We need to note that although \textbf{\ours} has 7B parameters in total, the activated parameters for each token are only 3B, indicating a lower inference cost than Janus-Pro-7B. Compared to the strong generation-only model, SD3-Medium~\cite{esser2024scaling}, \textbf{\ours} also achieves better overall performance, especially on the scores of position. Results on DPG-Bench~\cite{dpg-bench} and GenAI-Bench~\cite{lin2024evaluating} are shown in Table~\ref{tab:exp-dpg} and Table~\ref{tab:exp-genai}, respectively. The conclusions on different benchmarks are consistent, showing the strong generation ability of \textbf{\ours}. 

\paragraph{\textbf{Human Evaluation}}
In order to better evaluate the model from the subjective perspective of humans, we have further conducted manual evaluations of {\ours} and other models on the Bench-240 proposed by Seedream 2.0~\cite{gong2025seedream20nativechineseenglish}. The evaluations are mainly carried out from two aspects: text-image alignment, and structural correction. The evaluations are conducted by expert reviewers, and the specific evaluation strategy is consistent with that in Seedream 2.0. Expert reviewers will rate each model on a scale from 1 (indicating extreme dissatisfaction) to 5 (indicating extreme satisfaction), and the final score of a model is the arithmetic mean of the scores given by multiple reviewers.

\begin{figure*}[h]
 \centering
 \includegraphics[width=0.6\linewidth]{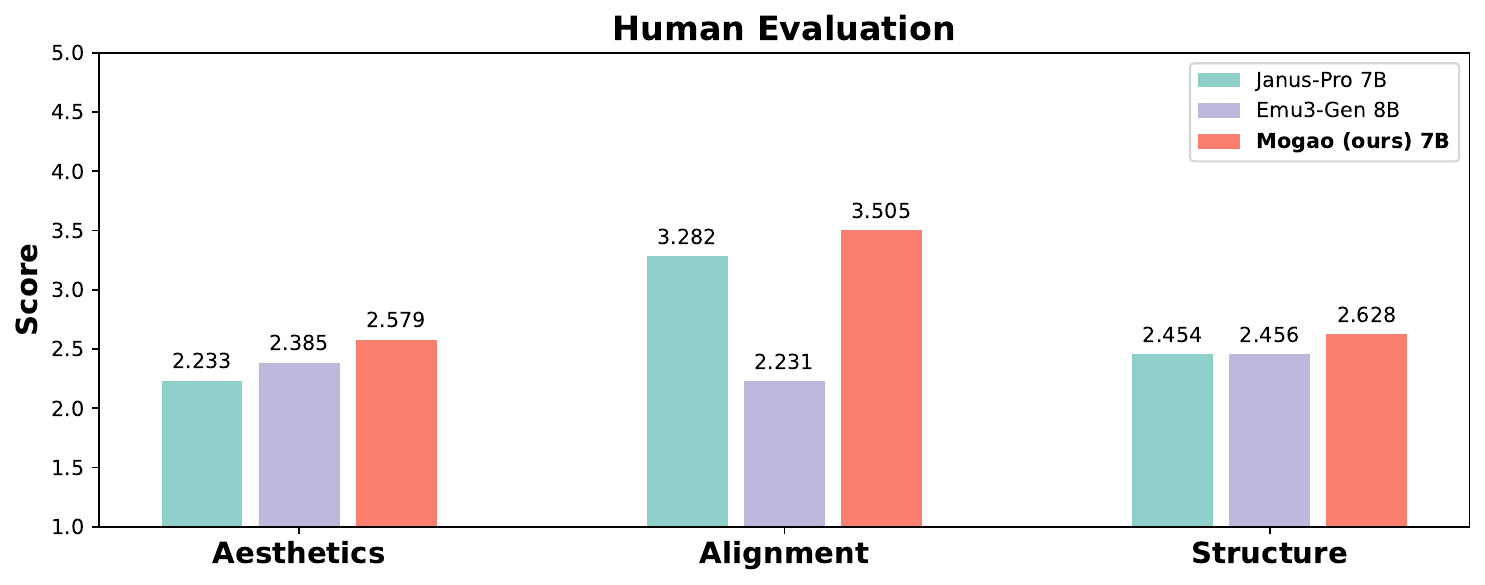}
 \caption{Human evaluation.}
 \label{fig:human_evaluation}
 \end{figure*}

The results in Fig.~\ref{fig:human_evaluation} demonstrate that {\ours} not only performs well on open-source benchmarks but also ranks first in human evaluations across three dimensions, significantly outperforming advanced unified models such as Emu3-Gen and Janus-Pro. This suggests that {\ours} has the potential to serve as a foundational model that is more practical in real-world scenarios. 

\paragraph{\textbf{Text-to-Image Visualization}}
Mogao demonstrates remarkable aesthetic performance in image generation, as illustrated in Figure 1. Despite being fine-tuned with a limited dataset of high-quality $512\times512$ images, the model excels in image quality and aesthetic presentation. The generated images are impressive in composition, color manipulation, reality-virtual interplay, and emotional conveyance. In addition, Mogao’s inherent text rendering capability allows it to generate poster-style images that push the boundaries of creativity.


\subsection{Multi-modal Understanding}
\input{mimo_sections/table/und_eval_table}

To evaluate {\ours}'s performance on multi-modal understanding, we show a comprehensive comparison of {\ours} with other methods on various public benchmarks in Tab.~\ref{tab:und}. The compared models can be divided into two categories: multi-modal understanding models and multi-modal unified models. {\ours} achieved SOTA on 4 benchmarks, including POPE, MME-P, SEED, and MMMU, and obtained the highest average score over 6 benchmarks. This result shows that {\ours} has well integrated the two tasks of image understanding and image generation into one model and achieved satisfying performance.

\subsection{Ablation Studies}

\input{mimo_sections/table/ablation}


 \begin{figure*}[h]
 \centering
 \includegraphics[width=0.5\linewidth]{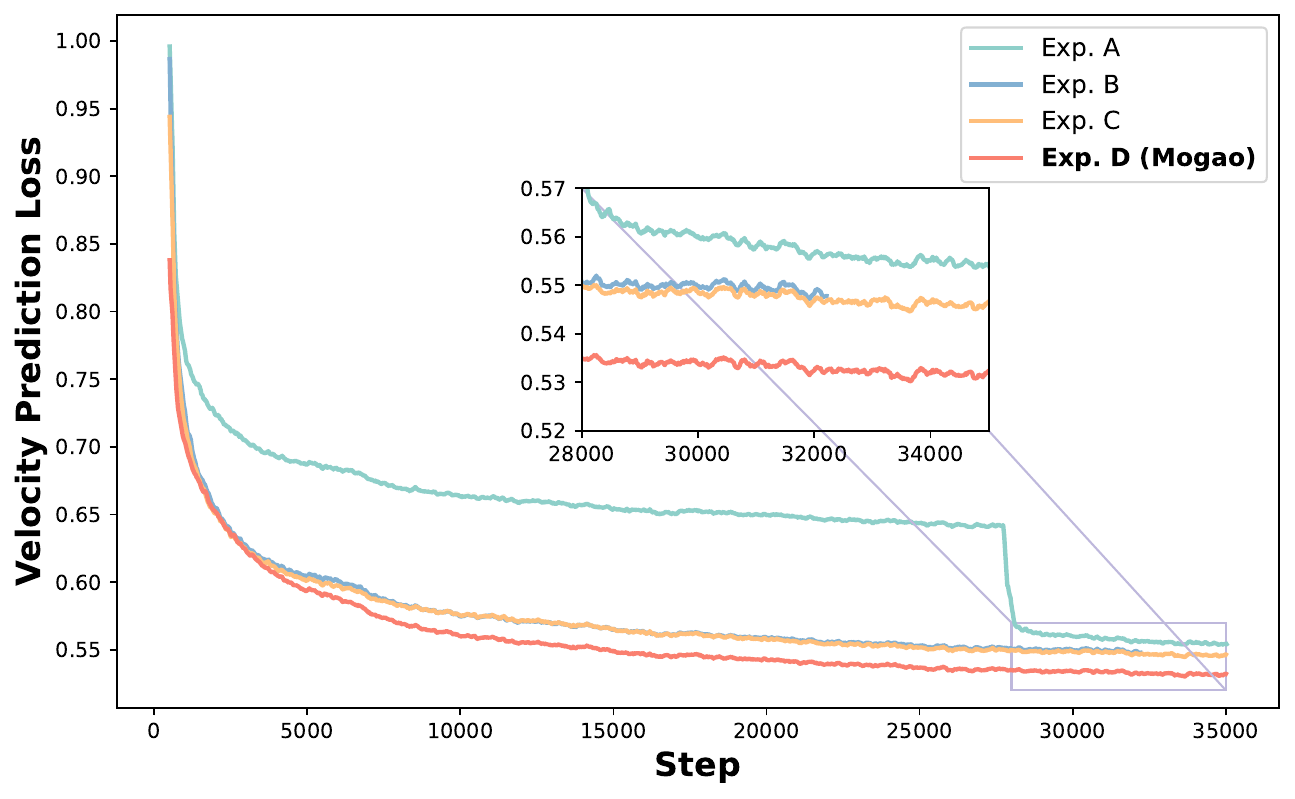}
 \caption{\textbf{Ablation studies on flow matching loss.} According to our experimental experience, a reduction in the velocity prediction loss of \textbf{0.001-level} will lead to a significant improvement in image generation.}
 \label{fig:abl_vloss}
 \end{figure*}

The qualitative and quantitative analysis in Sec.~\ref{sec:exp} demonstrates that {\ours} is an omni-foundation model with interleaved multi-modal generation capabilities. In this section, we conducted comprehensive ablation experiments to verify the effectiveness and necessity of the key designs in Sec.~\ref{sec:method}. All ablation experiments will be conducted in a development setting. Concretely, the maximum image resolution is set to $256 \times 256$, the 1.5B model of Qwen2 is used as the backbone, and all models are trained for 35,000 steps. The settings of different experiments and the multi-modal understanding metrics are shown in Tab.~\ref{tab:ablation}, and the loss curves of image generation are shown in Fig.~\ref{fig:abl_vloss}.

\paragraph{\textbf{Position Embedding for Unified Model}}
 As shown in Fig.~\ref{fig:abl_vloss}, the comparison between Exp.~A and Exp.~B demonstrates that the IL-RoPE proposed in this paper has a significant advantage in image generation, and its loss converges faster and is lower. We attribute this to the reasonable frequency allocation detailed in Sec.~\ref{sec:method_model}. In contrast, the unreasonable frequency allocation strategy of M-RoPE in the $H$ and $W$ dimensions leads to a slower loss convergence speed.

\paragraph{\textbf{Enhance Multi-Modal Understanding with Dual Visual Encoder}} 
 Comparing Exp.~B and Exp.~C in Tab.~\ref{tab:ablation}, we can find that the image representation with high-level semantics from ViT significantly enhances {\ours}'s multi-modal understanding ability. Furthermore, we can observe that Exp.~C has a lower image generation loss in Fig.~\ref{fig:abl_vloss}. This is because diffusion focuses more on modeling low-frequency semantics and image structure in the high-noise stage, while the additional visual representation from ViT added during training can help the model better learn high-level semantics. Similar findings can be found in REPA~\cite{yu2024representation}.

\paragraph{\textbf{Deep-Fusion Architecture}} 
 In Exp.~C, the image and text modalities share parameters in the backbone, while Exp.~D adopts the deep fusion architecture described in Sec.~\ref{sec:method_model}. As depicted in Fig.~\ref{fig:abl_vloss} and Tab.~\ref{tab:ablation}, the flow matching loss of Exp.~D is significantly reduced compared to Exp.~C, and the average score over 6 multi-modal understanding benchmarks is further improved. We attribute this phenomenon to the fact that multi-modal understanding tasks favor low-frequency, high-level semantic representations, while generation tasks, particularly in the low-noise stage, prioritize high-frequency, low-level detail representations~\cite{rissanen2022generative}. Thus, decoupling understanding and generation within the backbone enhances the performance of both tasks effectively.

%% file: mimo_sections/table/t2i_eval_table.tex
 \begin{table}[!h]
    \centering
    
    \caption{\textbf{Evaluation of text-to-image generation ability on GenEval benchmark.}}\label{tab:geneval}
    
    \setlength{\tabcolsep}{3mm}
    \resizebox{0.8\linewidth}{!}{
    \begin{tabular}{lccccccc}
        \toprule
        \textbf{Method}  & \textbf{Single Obj.} & \textbf{Two Obj.} & \textbf{Counting} & \textbf{Colors} & \textbf{Position} & \textbf{Color Attri.} & \textbf{Overall$\uparrow$} \\
        \midrule
        \multicolumn{8}{l}{\textit{Multi-Modal Generation Models}}\\
        \midrule
        LlamaGen~\cite{llamagen}  & $0.71$ & $0.34$ & $0.21$ & $0.58$ & $0.07$ & $0.04$ & $0.32$ \\
        LDM~\cite{ldm} & $0.92$ & $0.29$ & $0.23$ & $0.70$ & $0.02$ & $0.05$ & $0.37$ \\
        SDv$1.5$~\cite{ldm} &  $0.97$ & $0.38$ & $0.35$ & $0.76$ & $0.04$ & $0.06$ & $0.43$ \\
        PixArt-$\alpha$~\cite{chen2023pixart} &  $0.98$ & $0.50$ & $0.44$ & $0.80$ & $0.08$ & $0.07$ & $0.48$ \\
        SDv$2.1$~\cite{ldm} &  $0.98$ & $0.51$ & $0.44$ & $0.85$ & $0.07$ & $0.17$ & $0.50$ \\
        DALL-E $2$~\cite{dalle2}  & $0.94$ & $0.66$ & $0.49$ & $0.77$ & $0.10$ & $0.19$ & $0.52$ \\
        Emu$3$-Gen ~\cite{wang2024emu3}  & $0.98$ & $0.71$ & $0.34$ & $0.81$ & $0.17$ & $0.21$ & $0.54$ \\
        SDXL~\cite{sdxl} &  $0.98$ & $0.74$ & $0.39$ & $0.85$ & $0.15$ & $0.23$ & $0.55$ \\
        DALL-E $3$~\cite{dalle3}  & $0.96$ & $0.87$ & $0.47$ & $0.83$ & $0.43$ & $0.45$ & $0.67$ \\
        SD3-Medium~\cite{esser2024scaling} & 0.99 & 0.94 & 0.72 & 0.89 & 0.33 & 0.60 & $0.74$ \\
        \midrule
        \multicolumn{8}{l}{\textit{Multi-Modal Unified Models}}\\
        \midrule
        SEED-X$^\dagger$~\cite{seedx}  & $0.97$ & $0.58$ & $0.26$ & $0.80$ & $0.19$ & $0.14$ & $0.49$ \\
        Show-o~\cite{xie2024show} &  $0.95$ & $0.52$ & $0.49$ & $0.82$ & $0.11$ & $0.28$ & $0.53$ \\
        D-DiT~\cite{li2024dual} &  $0.97$ & $0.80$ & $0.54$ & $0.76$ & $0.32$ & $0.50$ & $0.65$ \\
        LWM~\cite{liu2024world} &  $0.93$ & $0.41$ & $0.46$ & $0.79$ & $0.09$ & $0.15$ & $0.47$ \\
        Transfusion~\cite{zhou2024transfusion} & - & - & - & - & - & - & $0.63$ \\
        ILLUME~\cite{wang2024illume} &  $0.99$ & $0.86$ & $0.45$ & $0.71$ & $0.39$ & $0.28$ & $0.61$ \\
        TokenFlow-XL~\cite{tokenflow} &  $0.95$ & $0.60$ & $0.41$ & $0.81$ & $0.16$ & $0.24$ & $0.55$ \\
        Chameleon~\cite{team2024chameleon} &  - & - & - & - & - & - & $0.39$ \\
        Janus~\cite{wu2024janus}
        & $0.97$ & $0.68$ & $0.30$ & $0.84$ & $0.46$ & $0.42$ & $0.61$ \\
        D-DiT~\cite{li2024dual}
        & $0.97$ & $0.80$ & $0.54$ & $0.76$ & $0.32$ & $0.50$ & $0.65$ \\
        Janus-Pro-1B~\cite{janus-pro} &  $0.98$ & $0.82$ & $0.51$ & $0.89$ & $0.65$ & $0.56$ & $0.73$ \\
        Janus-Pro-7B~\cite{janus-pro} &  $0.99$ & $0.89$ & $0.59$ & $0.90$ & $0.79$ & $0.66$ & $0.80$ \\
        \textbf{\ours-7B}&  $1.00$ & $0.97$ & $0.83$ & $0.93$ & $0.84$ & $0.80$ & $0.89$ \\
        \bottomrule
    \end{tabular}
}
    
\end{table}

\begin{table}[h]
    \centering
    \caption{\textbf{Performances on DPG-Bench.}}\label{tab:exp-dpg}
    
    \setlength{\tabcolsep}{5mm}
    \resizebox{0.8\linewidth}{!}{
    \begin{tabular}{lcccccc}
        \toprule
        \textbf{Method} & \textbf{Global} & \textbf{Entity} & \textbf{Attribute} & \textbf{Relation} & \textbf{Other} & \textbf{Overall$\uparrow$} \\
        \midrule
        \multicolumn{7}{l}{\textit{Multi-Modal Generation Models}} \\
        \midrule
        SDv1.5 \cite{ldm} 
        & 74.63 & 74.23 & 75.39 & 73.49 & 67.81 & 63.18 \\
        PixArt-$\alpha$ \cite{chen2023pixart} 
        & 74.97 & 79.32 & 78.60 & 82.57 & 76.96 & 71.11 \\
        Lumina-Next \cite{2024lumina} 
        & 82.82 & 88.65 & 86.44 & 80.53 & 81.82 & 74.63 \\
        SDXL \cite{sdxl} 
        & 83.27 & 82.43 & 80.91 & 86.76 & 80.41 & 74.65 \\
        Playground v2.5 \cite{2024PG2.5} 
        & 83.06 & 82.59 & 81.20 & 84.08 & 83.50 & 75.47 \\
        Hunyuan-DiT \cite{2024hunyuandit}
        & 84.59 & 80.59 & 88.01 & 74.36 & 86.41 & 78.87 \\
        PixArt-$\Sigma$ \cite{2024pixartsigma} 
        & 86.89 & 82.89 & 88.94 & 86.59 & 87.68 & 80.54\\
        Emu3-Gen \cite{wang2024emu3} 
        & 85.21 & 86.68 & 86.84 & 90.22 & 83.15 & 80.60 \\
        DALL-E 3~\cite{dalle3} 
        & 90.97 & 89.61 & 88.39 & 90.58 & 89.83 & 83.50 \\
        SD3-Medium \cite{esser2024scaling} 
        & 87.90 & 91.01 & 88.83 & 80.70 & 88.68 & 84.08 \\
        \midrule
        \multicolumn{7}{l}{\textit{Multi-Modal Unified Models}}\\
        \midrule
        Janus~\cite{wu2024janus} & 82.33 & 87.38 & 87.70 & 85.46 & 86.41 & 79.68 \\
        Janus-Pro-1B~\cite{janus-pro}
        & 87.58 & 88.63 & 88.17 & 88.98 & 88.30 & 82.63 \\
        Janus-Pro-7B~\cite{janus-pro}
        & 86.90 & 88.90 & 89.40 & 89.32 & 89.48 & 84.19 \\
        \textbf{\ours-7B}
        & 82.37 & 90.03 & 88.26 & 93.18 & 85.40 & 84.33 \\
        \bottomrule
    \end{tabular}}
\end{table}

\begin{table}[h]
\centering
\caption{\textbf{Comparisons on GenAI-Bench.} The results are reported on the advanced prompts of GenAI-Bench.}\label{tab:exp-genai}
\setlength{\tabcolsep}{5mm}
\resizebox{0.8\linewidth}{!}{
\begin{tabular}{lcccccc}
\toprule
\multirow{2}{*}{Methods} & \multirow{2}{*}{Count$\uparrow$} & \multirow{2}{*}{Differ$\uparrow$} & \multirow{2}{*}{Compare$\uparrow$} & \multicolumn{2}{c}{Logical$\uparrow$} & \multirow{2}{*}{Overall$\uparrow$} \\
\cmidrule{5-6}
& & & & Negate & Universal & \\
\midrule
\multicolumn{7}{l}{\textit{Multi-Modal Generation Models}}\\
\midrule
SDv2.1~\cite{ldm} & 0.68 & 0.70 & 0.68 & 0.54 & 0.64 & 0.62 \\
SD-XL~\cite{sdxl} & 0.71 & 0.73 & 0.69 & 0.50 & 0.66 & 0.63 \\
Mid-journey v6~\cite{mjv6} & 0.78 & 0.78 & 0.79 & 0.50 & 0.76 & 0.69 \\
DALL-E 3~\cite{dalle3} & 0.82 & 0.78 & 0.82 & 0.48 & 0.80 & 0.70 \\
\midrule
\multicolumn{7}{l}{\textit{Multi-Modal Unified Models}}\\
\midrule
show-o~\cite{xie2024show} & 0.70 & 0.62 & 0.71 & 0.51 & 0.65 & 0.60 \\
LWM~\cite{liu2024world} & 0.59 & 0.58 & 0.54 & 0.49 & 0.52 & 0.53 \\
VILA-U~\cite{vila-u} & 0.70 & 0.71 & 0.74 & 0.53 & 0.66 & 0.64 \\
Liquid~\cite{liquid} & 0.76 & 0.73 & 0.74 & 0.46 & 0.74 & 0.65 \\
UniTok~\cite{unitok} & 0.76 & 0.76 & 0.79 & 0.46 & 0.73 & 0.67 \\
 \textbf{\ours-7B} & 0.77 & 0.74 & 0.77 & 0.53 & 0.71 & 0.68 \\
\bottomrule
\end{tabular}}
\end{table}

%% file: mimo_sections/table/und_eval_table.tex
\begin{table}[ht]
    \centering
    
    \caption{\textbf{Comparison with other methods on multimodal understanding benchmarks}. "Avg." represents the average score over 6 benchmarks.
    }\label{tab:und}
    
    \setlength{\tabcolsep}{3mm}
    \resizebox{0.8\linewidth}{!}{
    \begin{tabular}{lccccccccc}
        \toprule
        \textbf{Model} & \textbf{Trainable Params} & \textbf{POPE} & \textbf{MME-P} & \textbf{MMB} & \textbf{SEED} & \textbf{GQA} & \textbf{MMMU} & \textbf{Avg.} \\
        \midrule
        \multicolumn{8}{l}{\textit{Multi-Modal Understanding Models}}\\
        \midrule
        MobileVLM-V2~\cite{chu2024mobilevlm2} & $1.4$B & $84.3$ & $1302.8$ & $57.7$ & - &  $59.3$ & - & -\\
        MobileVLM-V2~\cite{chu2024mobilevlm2} & $2.7$B & $84.7$ & $1440.5$ & $63.2$ & - &  $61.1$ & - & -\\
        LLaVA-Phi~\cite{zhu2024llava} & $2.7$B & $85.0$ & $1335.1$ & $59.8$ & -  & - & - &  - \\
        LLaVA-v$1.5$~\cite{liu2024improved}& $7$B & $85.9$ & $1510.7$ & $64.3$ & $58.6$ & $62.0$ & $35.4$ & - \\
        Qwen-VL-Chat~\cite{bai2023qwen} & $7$B & - & $1487.5$ & $60.6$ & $58.2$ & $57.5$ & - & - \\
        IDEFICS~\cite{laurencon2023introducing} & $8$B & - & - & $48.2$ & - &  $38.4$ & - & - \\
        Emu$3$-Chat~\cite{wang2024emu3} & $8$B & $85.2$ & $1244$ & $58.5$ & $68.2$ &  $60.3$ & $31.6$ & $50.7$ \\
        \midrule
        \multicolumn{8}{l}{\textit{Multi-Modal Unified Models}}\\
        \midrule
        Show-o-512~\cite{xie2024show} & $1.3$B & $80.0$ & $1097.2$ & - & - &  $58.0$ & $26.7$ & - \\
        JanusFlow \cite{ma2024janusflow} & 1.3B & 88.0 & 1333.1 & 74.9 & 70.5 & 60.3 & 29.3 & $53.9$ \\
        Janus-Pro \cite{janus-pro} & $1.5$B & $86.2$ & $1444.0$ & $75.5$ & $68.3$ & $59.3$ & $36.3$ & $54.4$ \\
        D-Dit~\cite{li2024dual} & $2$B & $84.0$ & $1124.7$ & - & - & $59.2$ & - & - \\
        ILLUME~\cite{wang2024illume} & $7$B &  $88.5$ &  $1445.3$ &  $65.1$ &  $72.9$ &   $-$ & $38.2$ &  - \\
        LWM~\cite{liu2024world} & $7$B & $75.2$ & - & - & - &  $44.8$ & - & - \\
        VILA-U~\cite{vila-u} & $7$B & $85.8$ & $1401.8$ & - & $59.0$ &  $60.8$ & - & - \\
        Janus-Pro~\cite{janus-pro} & $7$B & $87.4$ & $1567.1$ & $\mathbf{79.2}$ & $72.1$ & $62.0$ & $41.0$ & $57.1$ \\
        MMAR~\cite{yang2024mmar} & $7$B & $83.0$ & $1393.9$ & $66.3$ & $64.5$ & - & - & - \\
        MetaMorph~\cite{tong2024metamorph} & $8$B & - & - & $75.2$ & $71.8$  & - & - & - \\
        TokenFlow-XL~\cite{tokenflow} & $13$B &  $86.8$ &  $1545.9$ &  $68.9$ &  $68.7$ &   $\mathbf{62.7}$ & $38.7$ &  $54.4$ \\
        \textbf{{\ours-7B} (ours)} & $7$B &  $\mathbf{88.9}$ &  $\mathbf{1592}$ &  75.0 &  $\mathbf{74.6}$ &   60.9 & $\mathbf{44.2}$ &  $\mathbf{57.4}$ \\
        \bottomrule
    \end{tabular}
    }
\end{table}

%% file: mimo_sections/table/ablation.tex
\begin{table}[!h]
    \centering
    \caption{\textbf{Ablation studies on multi-modal understanding benchmark.} Experiment with \underline{underline} denote the baseline settings and {\ours} setting is highlighted in \colorbox{gray!30}{gray}.}
    \label{tab:ablation}
    \setlength{\tabcolsep}{2mm}
    \resizebox{0.85\linewidth}{!}{
    \begin{tabular}{ccccccccccccc}
        \toprule
        \multirow{2}{*}{\textbf{Exp.~ID}} & \multirow{2}{*}{\textbf{Deep-Fusion}} & \multirow{2}{*}{\textbf{Position Embedding}} & \multicolumn{2}{c}{\textbf{Visual Encoder}} & \multicolumn{6}{c}{\textbf{Evaluation Benchmarks}} \\
                                          & &  & \textbf{Gen.} & \textbf{Und.}  & \textbf{POPE} & \textbf{MME-P} & \textbf{MMB} & \textbf{SEED} & \textbf{GQA} & \textbf{MMMU} & \textbf{Avg.}\\
        \midrule
        \underline{A} & $\times$     & M-RoPE~\cite{qwen2.5-vl} & VAE  & VAE & $67.2$ & $885.9$ & $29.9$ & $48.2$ & $46.7$ & $32.8$ & $37.5$ \\
        \midrule
        B & $\times$     & IL-RoPE  & VAE & VAE & $76.4$ & $880.3$ & $29.2$ & $51.8$ & $47.2$ & $33.3$ & $39.7$ \\
        \midrule
        C & $\times$     & IL-RoPE  & VAE & VAE + ViT & $86.6$ & $1302.6$ & $62.5$ & $\mathbf{68.4}$ & $56.9$ & $35.3$ & $51.7$ \\
        \midrule
        \rowcolor{gray!20} D \textbf{(\ours)} & $\checkmark$ & IL-RoPE  & VAE & VAE + ViT & $\mathbf{87.3}$ & $\mathbf{1353.6}$ & $\mathbf{65.9}$ & $66.8$ & $\mathbf{58.3}$ & $\mathbf{36.7}$ & $\mathbf{52.6}$ \\
        \bottomrule
    \end{tabular}
    }
\end{table}

%% file: mimo_sections/conclusion.tex
\section{Conclusion}
We introduce {\ours}, an omni-foundation model for interleaved multi-modal generation. We first propose several key model architecture designs to seamlessly integrate autoregressive and diffusion models in {\ours}, which enable {\ours} to achieve excellent multi-modal understanding and image generation capabilities. Based on the strong single-modal generation capabilities, {\ours} unlocks more powerful interleaved multi-modal generation capabilities than previous unified models by performing causal interleaved generation at the modal level. Through an efficient large-scale training process utilizing interleaved multi-modal data, {\ours} not only excels in multi-modal generation tasks but also exhibits competitive emergent capabilities, such as zero-shot image editing and element composition, with strong consistency. Overall, unified models demonstrate greater potential compared to single-modal generation approaches. We hope that {\ours} will inspire further exploration to advance the development of general artificial intelligence.